\documentclass{article}

\usepackage{arxiv}
\usepackage[utf8]{inputenc} 
\usepackage[T1]{fontenc}    
\usepackage{hyperref}       
\usepackage{url}            
\usepackage{booktabs}       
\usepackage{amsfonts}       
\usepackage{nicefrac}       
\usepackage{microtype}      
\usepackage{lipsum}
\usepackage{graphicx}

\usepackage{times}    
\usepackage{caption}
\usepackage{latexsym}
\usepackage[misc]{ifsym}
\usepackage{amsmath}
\usepackage{amssymb}
\usepackage{bbm}
\usepackage[ruled,vlined,linesnumbered,noresetcount]{algorithm2e}
\usepackage{array}
\newcolumntype{P}[1]{>{\centering\arraybackslash}p{#1}}
\usepackage[caption=false]{subfig}

\graphicspath{ {./images/} }

\title{COLAM: Co-Learning of Deep Neural Networks and Soft Labels via Alternating Minimization}

\author{
  Xingjian Li\footnotemark[1] \\
  Big Data Lab, Baidu Research\\
  University of Macau\\
  \texttt{lixingjian@baidu.com} \\
   \And
  Haoyi Xiong\footnotemark[1] \\
  Big Data Lab, Baidu Research\\
  \texttt{xionghaoyi@baidu.com} \\
  \And
 Haozhe An \\
  Big Data Lab, Baidu Research\\
  \texttt{v\_anhaozhe@baidu.com} \\
  \And
 Dejing Dou \\
  Big Data Lab, Baidu Research\\
  \texttt{doudejing@baidu.com} \\
  \And
  Chengzhong Xu \\
  University of Macau\\
  \texttt{czxu@um.edu.mo} \\
}
\begin{document}
\maketitle

\renewcommand{\thefootnote}{\fnsymbol{footnote}}
\footnotetext[1]{These authors contributed equally to this work.} 

\begin{abstract}
Softening labels of training datasets with respect to data representations has been frequently used to improve the training of deep neural networks (DNNs). While such a practice has been studied as a way to leverage  ``privileged information'' about the distribution of the data, a well-trained learner with soft classification outputs should be first obtained as a prior to generate such privileged information. To solve such a \emph{``chicken-and-egg''} problem, we propose COLAM framework that \underline{C}o-\underline{L}earns DNNs and soft labels through \underline{A}lternating \underline{M}inimization of two objectives -- (a) the training loss subject to soft labels and (b) the objective to learn improved soft labels -- in one end-to-end training procedure. We performed extensive experiments to compare our proposed method with a series of baselines. The experiment results show that COLAM achieves improved performance on many tasks with better testing classification accuracy. We also provide both qualitative and quantitative analyses that explain why COLAM works well.
\end{abstract}

\keywords{deep learning, neural networks, soft label}

\maketitle

\section{Introduction}
Recent years have witnessed rapid developments in deep neural networks~\cite{He2015DeepRL,Simonyan2014VeryDC,Szegedy2014GoingDW,szegedy2016rethinking} and their widespread applications in a variety of tasks~\cite{Graves2014TowardsES,Jzefowicz2016ExploringTL}. Due to the over-parameterized nature of deep neural networks~\cite{Pereyra2017RegularizingNN,Zhang2016UnderstandingDL}, tons of tricks have been invented to improve the generalization performance of deep learning through regularizing the training procedure, such as weigh decay~\cite{Krizhevsky2009LearningML,Zhang2018ThreeMO}, DropOut~\cite{Hinton2012ImprovingNN}, stochastic pooling~\cite{Zeiler2013StochasticPF}, data augmentation~\cite{Krizhevsky2012ImageNetCW}, as well as using perturbed labels~\cite{Xie2016DisturbLabelRC} or soft labels~\cite{szegedy2016rethinking} (drawn from the continuous space) to replace the original hard labels (zero-one coded) of training samples. In this work, we study the problem of learning optimal soft labels for training samples subject to the deep learning process, and further propose novel algorithm that intends to co-learn both ``best'' soft labels and deep neural networks through an end-to-end training procedure.

Researchers have widely adopted label smoothing to optimize a broad range of tasks, including but not limited to image classification~\cite{Huang2018GPipeET,Real2018RegularizedEF,Zoph2017LearningTA}, speech recognition~\cite{Chorowski2016TowardsBD}, and machine translation~\cite{Vaswani2017AttentionIA}. The key principle here to regularize the deep learning procedure with certain privileged prior information~\cite{Xie2016DisturbLabelRC,lopez2016unifying} embedded in the soft labels. %
With a set of predefined rules, label smoothing~\cite{szegedy2016rethinking} was first proposed to soften the hard labels to regularize the training objectives with smoothness. In addition to using predefined mappings, learning from the soft classification outputs (e.g., logits) of a well-trained teacher neural network (often named as knowledge distillation~\cite{hinton2015distilling}) could also improve the generalization performance significantly. In our research, we soften labels using well-trained neural networks, so as to incorporate the privileged knowledge of teacher network~\cite{lopez2016unifying}. More specifically, due to the lack of well-trained models in advance, the proposed algorithm is expected to learn optimal soft labels from the DNN outputs during the training process, i.e., under self-distillation settings. 

To achieve the goal, we propose a novel deep learning algorithm, namely COLAM -- the \underline{CO}-\underline{L}earning of deep neural networks and soft labels via \underline{A}lternating \underline{M}inimization. During the training procedure, COLAM alternatively minimizes two learning objectives: (i) the training loss subject to the target (soft) labels, and (ii) the loss to learn soft label design subject to the logit outputs of learned labels. Compared to the existing solution that either use the raw soft prediction results  as the soft labels for self-distillation, or leverage pre-trained models  as teacher networks with additional computation cost required, COLAM uses one end-to-end training procedure to effectively learn both soft labels (of all training samples) and the model all in once. COLAM improves the generalization of deep learning through softening the labels with ``privileged'' information while enjoying the same computation cost of vanilla training.

The contributions made in this paper are as follows. We study the technical problem of co-learning soft labels and deep neural network  during one end-to-end training process in a self-distillation setting. We design two objectives to learn the model and the soft labels respectively, where the two objective functions depend on each other.  We further propose COLAM algorithm that achieves the goal through alternatively minimizing two objectives. Extensive  experiments have been done using real-world image classification datasets under the both supervised learning and transfer learning settings. We compare COLAM with a bunch of baselines algorithms using soft labels and perturbed labels. The experiment results showed that COLAM can significantly outperform baseline methods with significantly higher classification accuracy (1\%--2\%) using comparable computation cost. 

\section{Related Work and Backgrounds}
Label smoothing (LS) was first introduced in~\cite{szegedy2016rethinking} to enhance the performance of Inception model on ImageNet~\cite{Deng2009ImageNetAL}. This traditional label smoothing is a weighted average of ground-truth label. Formally, given a ground-truth label $\pmb{y} = (\pmb{y}^1, \pmb{y}^2, \dots, \pmb{y}^n)$ in a classification task for $n$ classes, we have $\forall i \in [1, n], \pmb{y}^i \in \{0, 1\}$ and $\sum_i \pmb{y}^i = 1$. If $\pmb{y}_l = 1$, then $l$ is the correct class to which that sample belongs. The softened label is obtained by
\begin{equation}
    \pmb{y}^i_{soft} = \pmb{y}^i(1-\epsilon) + \epsilon/n.
\end{equation}
Note that the superscripts represent the indices. The hard label is replaced by the softened label when computing the cross entropy loss between label and predicted probabilities. Mathematically, the cross entropy between the ground-truth targets $y$ and a predicted probability distribution $p$ is 
\begin{equation}
    H(\pmb{y},p) = \sum_i -\pmb{y}^i \log (p^i).
\end{equation}
Now the soft label substitutes the ground-truth hard label in the cross entropy function, giving rise to
\begin{equation}
    H(\pmb{y},p) = \sum_i -\pmb{y}_{soft}^i \log (p_i).
\end{equation}
This noisy loss result enables the network to reduce the chances of being overconfident while making predictions, thus regularizing the network.
Besides following a uniformed distribution to produce soft labels, label smoothing can be more dynamic. \cite{Pereyra2017RegularizingNN} shows that, with a slight modification of the KL Divergence direction, ``confidence penalty" regularizer is equivalent to label smoothing. This regularizer encourages predictions to have larger entropy and lower confidence on the most probable class. It is achieved by softening the model output. 

Furthermore, adding noise to labels produces similar effects as label smoothing. DisturbLabel~\cite{Xie2016DisturbLabelRC} is a regularizer in the loss layer of a network. It adds noise to labels during training by randomly changing a correct label $\pmb{y}$ to another one-hot label $\Tilde{\pmb{y}}$. This permutation of the elements in labels happens under a certain fixed likelihood. \cite{Xie2016DisturbLabelRC} points out DisturbLabel has the same expected gradient as label smoothing because $\mathbbm{E}(\Tilde{\pmb{y}}) = \pmb{y}_{soft}$. However, DisturbLabel outperforms traditional LS on many datasets, likely because randomness in the algorithm contributes to the success. Although these regularizers bring improvements in generalization, little to no dataset knowledge is involved to produce soft labels.

\begin{algorithm} 
 \KwIn{Initial parameters $\pmb\theta_{M_0}$; the temperature $T$; and the set of training samples $\mathcal{D}$}
 \KwOut{Deep neural network model parameters $\pmb{\theta}_{M_m}$}
    \For{$n=1,\ 2,\ 3,\dots,\ m$}{
        \text{/* Training from $\pmb{\theta}_{M_{n-1}}$ with $t$ epochs */}\\
        \eIf{$M_n$ is the first stage $M_1$}{
        \text{/* Using Original Hard Labels  */}\\
            $\pmb{\theta}_{M_n}\gets\mathrm{arg min}_\theta\ \sum_{i=1}^{|\mathcal{D}|} \mathcal{L}(\theta; (\pmb{x}_i,y_i),T)$
        }{
                \text{/* Using (Updated) Soft Labels */}\\
            $\pmb{\theta}_{M_n}\gets\mathrm{arg min}_\theta\ \sum_{i=1}^{|\mathcal{D}|} \mathcal{L}(\theta; (\pmb{x}_i,\pmb{y}_{i,\mathrm{soft}}),T)$
        }
\text{/* Updating the Soft Labels using $\pmb{\theta}_{M_n}$*/}\\
    \For {$i=1,\ 2,\ 3, \dots, |\mathcal{D}|$} {
    \text{Obtaining peer samples of $(\pmb{x}_i,y_i)$}\\
        $Y_{y_i}\gets\mathrm{getPeerSamples}(\pmb{x}_i,y_i);$\\
        $\pmb{y}'_i\gets\underset{\forall y\in\mathbb{R}^{|C|}}{\mathrm{arg\ min}}\underset{{\forall (x,y)\in Y_{y_i}}}{\sum} \mathrm{dist}(y;f(x;\pmb{\theta}_{M_n}))$\\
        $\pmb{y}_i\gets\mathrm{softmax}(\pmb{y}'_i/T)$
    }
	\Return{$\pmb{\theta}_{M_m}$}    
    }
 \caption{COLAM Algorithm}
 \label{implement}
\end{algorithm}

\section{COLAM Algorithm Design}
In this section, we first present the overall learning procedure with the design of two objectives.
%


\subsection{Learning Procedure}
 Let $\mathcal{D} \in \mathcal{X} \times \mathcal{Y}$ be the training set which contains $|\mathcal{D}|$ labelled training samples and $C$ classes. Each sample is denoted by $(\mathbf{x_i},y_i) \in \mathcal{D}$, where $y_i \in \{1, 2, \dots, C\}$. We define our objective deep neural network network $\pmb{f}(\mathbf{x};\pmb{\theta}): \mathcal{X} \mapsto \mathcal{Y}$ parameterized with $\pmb{\theta}$ as the mapping function. 

COLAM splits the overall training procedure into $m$ equal-length stages $M = \{M_1, M_2, \dots, M_m\}$, with each stage consisting of $T$ epochs. In the first stage, COLAM uses the original hard labels of training samples to train deep neural network with $t$ epochs and obtains $\theta_{M1}$. Then, COLAM computes the soft label for every sample (i.e., $\pmb{y}_{i,soft}$ for sample $(\pmb{x}_i,y_i)$) in the training set through minimizing the \emph{Loss of Soft Label Learning}. 

From the second stage to the final stage of the training procedure, COLAM continues the deep learning procedure using the training samples with soft labels ($(\pmb{x}_i,\pmb{y}_{i,soft})$) via minimizing the \emph{Loss of Model Learning}, and repeats the soft label computation by the end of stage. Through alternatively minimizing the training loss and soft label design loss, COLAM is expected to reach the convergence of deep learning and outputs the $\pmb{\theta}_{M_m}$ as the results of soft label and model co-learning. The model would be trained using $m\times t$ epochs.

The complete algorithm is illustrated in Algorithm~\ref{implement}.


\subsection{Loss of Deep Neural Network Training}
COLAM simply uses the cross-entropy function as the loss to train deep neural networks. For the first stage, COLAM computes the DNN training loss using hard labels, while it starts leveraging the soft labels from the second stage. Given a pair of predictor and label $(x,y)$, the parameter $\theta$ and the temperature $T$ for softmax, COLAM considers the cross-entropy loss as follow.
\begin{equation}
     \mathcal{L}(\theta;(x,y),T) = - \sum_{j=1}^{C} y^j \log \left(\mathrm{softmax}^j(f(x;\theta)/T)\right),
     \label{loss}
\end{equation}
where $y^j$ refers to the $j^{th}$ dimension of the label $y$ and $\mathrm{sofmax}^j(\cdot)$ is the $j^{th}$ dimension of the input. The softmax function is defined as follow.
\begin{equation}
   \mathrm{softmax}^j(y) = \frac{\exp (y^j)}{\sum_{c\in \{1,...C\}\setminus \{j\}} \exp (y^c) },\ \forall 1\leq j\leq C.
\end{equation}
 Note that, for the first stage, COLAM uses $(x,y)\in\mathcal{D}$ referring to the training sample with hard labels. From the second stage, COLAM uses $(x,y)\in\{(\pmb{x}_i,\pmb{y}_{i,\mathrm{soft}})|\forall 1\leq i\leq |\mathcal{D}|\}$ referring to the sample with the soft labels. Please refer to lines 3--8 of Algorithm.~1 for details.


\subsection{Loss of Soft Label Learning}
To achieve the better generalization while lowering the computation complexity, COLAM assumes all samples of the same class share the same soft label. 

\textbf{Peer Samples}. In this way, to learn the soft label, COLAM first retrieves the peer samples for every training sample.  Given $\mathcal{D}_i = (\mathbf{x_i}, y_i)$,  its peer sample set (denoted as $Y_{y^i}$) is defined as $Y_{y^i} =  \big\{ \forall (\mathbf{x}, y) \in \mathcal{D} ~|~ y=y_i\big\}$.

\textbf{Soft Label Loss}. Given the set of peer samples $Y_{y^i}$ for the class $y_i$, COLAM computes the soft label  $\pmb{y}'_i$ through minimizing the distances between the soft label to the soft prediction results of every sample in $Y_{y^i}$ (please see also in Line 13 of Algorithm.~1). Such that, COLAM simply defines the distance as follow.
\begin{equation}
\mathrm{dist}(y;y')=\frac{1}{2}\|y-y'\|_2^2,
\end{equation}
where $y'$ refers to the soft prediction result and $y$ is the learning objective of soft labels. To simplify the computation, line 13 of Algorithm 1 is equivalent to estimate the mean soft labels among all peer samples.  With the $\pmb{y}'_i$ obtained, COLAM uses softmax to further normalize the vector. Finally, COLAM uses $\pmb{y}_i$ to replace $y_i$ as the soft label for further computation. Please refer to lines 9--14 of Algorithm.~1 for details.

\begin{table*}
\caption{\textbf{Test accuracy on CIFAR10.} HL refers to models trained with standard Stochastic Gradient Descent optimizer using hard labels. LS are models improved by using Label Smoothing technique. CP refers to models regularized by Confidence Penalty. DL refers to DisturbLabel.}
\label{cifar10}
\begin{center}
\begin{tabular}{P{3cm}P{2.2cm}P{1.5cm}P{1.5cm}P{1.5cm}P{1.5cm}P{1.5cm}}
\toprule
\textbf{Model} & \textbf{\# of Parameters} & \textbf{HL} &\textbf{LS}& \textbf{CP} & \textbf{DL} &  \textbf{COLAM} \\
\midrule
DenseNet-BC-100 & 1M & 0.9474  & 0.9481& \textbf{0.9495} & 0.9481 & 0.9484\\ 
WideResNet40x4 & 9M & 0.9559 & 0.9558 & 0.9551 & 0.9566 & \textbf{0.9569} \\ 
PreActResNet34 & 21M & 0.9502 & 0.9501 & 	0.9512  &	0.9519 &\textbf{0.9521}\\ 
ResNeXt29\_8x64d & 34M & 0.9482	& 0.9512 & 0.9579 & 0.9552 & \textbf{0.9591} \\ 
\bottomrule
\end{tabular}
\end{center}
\end{table*}
\begin{table*}
\caption{\textbf{Test accuracy on CIFAR100.} Same abbreviations are used as found in Table.~\ref{cifar10}}
\label{cifar100}
\begin{center}
\begin{tabular}{P{3cm}P{2.2cm}P{1.5cm}P{1.5cm}P{1.5cm}P{1.5cm}P{1.5cm}}
\toprule
\textbf{Model} & \textbf{\# of Parameters} & \textbf{HL} &\textbf{LS} & \textbf{CP} & \textbf{DL} &  \textbf{COLAM} \\
\midrule
DenseNet-BC-100 & 1M & 0.7515 &	0.7551 &0.7543 & 0.7591 &	\textbf{0.7615}\\ 
WideResNet40x4 & 9M & 0.7799 & 0.7808& 	0.7781 & 0.7845 &\textbf{0.7868} \\ 
PreActResNet34 & 21M & 0.7701 & 0.7827& 0.7755 & 0.7762 &\textbf{0.7869} \\ 
ResNeXt29\_8x64d & 34M & 0.8017	& 0.8103 & 0.803 & 0.8064 & \textbf{0.8147} \\ 
\bottomrule
\end{tabular}
\end{center}
\end{table*}

\begin{table*}
\caption{\textbf{Test accuracy on ImageNet.} Same abbreviations are used as found in Table.~\ref{cifar10}.}
\label{imagenet}
\begin{center}
\begin{tabular}{P{3cm}P{2.2cm}P{1.5cm}P{1.5cm}P{1.5cm}P{1.5cm}P{1.5cm}}
\toprule
\textbf{Model} & \textbf{\# of Parameters} & \textbf{HL} &\textbf{LS} & \textbf{CP} & \textbf{DL} &  \textbf{COLAM} \\
\midrule
ResNet50  & 26M & 0.7658 &	0.7669 &0.7665 & 0.768 &	\textbf{0.7756}\\ 
ResNet101 & 45M & 0.7786 & 0.7841& 	0.7812   &	0.7833 &\textbf{0.7875} \\
\bottomrule
\end{tabular}
\end{center}
\end{table*}

\section{Experiments}

\subsection{Tasks and Datasets}
\noindent\textbf{Image Classification} We use CIFAR10, CIFAR100~\cite{Krizhevsky2009LearningML} and ImageNet~\cite{Deng2009ImageNetAL} to test the performance of COLAM on image classification task. CIFAR10 has 10 different classes. In the training set of CIFAR10, each class contains 5,000 images. CIFAR 100 contains RGB images categorized into 100 classes, with each class composing 600 images. There are 500 training images and 100 testing images in each class. ImageNet is a tree-structured image database created according to the WordNet hierarchy. It consists of more than 20K categories and a total of 14 million images. We use the popular subset ILSVRC2012 which contains 1.3M images covering 1K categories.

\noindent\textbf{Transfer Learning} We use ImageNet as the source dataset and 4 different target tasks covering typical types of plants, animals, objects and texture . The 4 target datasets are (1) Flower102~\cite{Nilsback2008AutomatedFC} which contains 102 categories of 8189 flower images, (2) Caltech-UCSD Birds-200-2011~\cite{WahCUB_200_2011}, which has 11,788 images classified into 200 categories, (3) FGVC-Aircraft~\cite{Maji2013FineGrainedVC} which composes 10,000 images of aircraft across 100 aircraft models, and (4) Describable Textures Dataset (DTD)~\cite{Cimpoi2015DeepFB} which is a texture database, consisting of 5640 images, organized according to a list of 47 terms (categories).
\label{data}

\subsection{Settings}
\noindent\textbf{Image Classification} We train PreActResNet34~\cite{He2016IdentityMI}, WideResNet40x4~\cite{Zagoruyko2016WideRN} and ResNeXt29\_8x64d~\cite{Xie2016AggregatedRT} for 160 epochs. We train DenseNet-BC-100~\cite{Huang2016DenselyCC} for 240 epochs because it converges slower. The initial learning rate is 0.1 for all architectures. Training batch size is 64. We use standard SGD optimizer with momentum 0.9 and weight decay 0.0001. We apply standard data augmentation the same way as the Pytorch official examples on CIFAR10, CIFAR100 and ImageNet classification task. For CIFAR dataset, we pad the input images by 4 pixels, and then randomly crop a sub-region of $32\times 32$ and randomly do a horizontal flip. For ImageNet, we first randomly crop a sub-region of $224\times 224$ and randomly do a horizontal flip. We normalize the input data as done in common practice.

\noindent\textbf{Transfer Learning} We use ResNet101~\cite{He2015DeepRL} as the base model to apply COLAM. We train the model with 40 epochs and the batch size for training is 64. SGD optimizer is used with a momentum of 0.9. The initial learning rate is set to 0.01 and the weight decay is set to 0.0001. We use exactly the same data augmentation methods as in ImageNet classification task. We repeat all these experiments 3 times and report the average Top-1 accuracy. 

\begin{figure*}
\begin{center}
   \subfloat[learning curve ]{\includegraphics[width=0.5\textwidth]{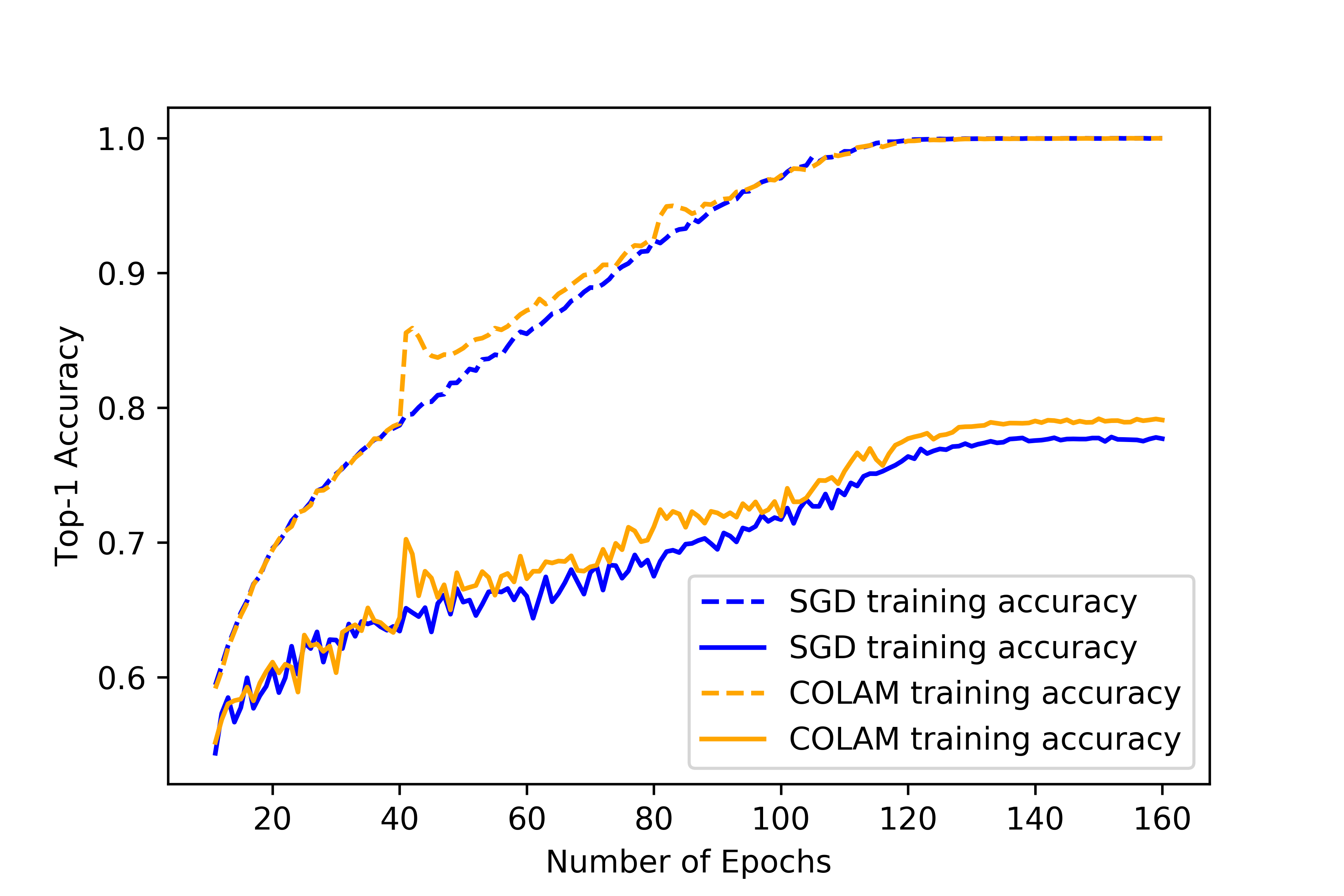}\label{learningcurve}}
   \subfloat[expected accuracy of the evolutionary soft labels]{\includegraphics[width=0.5\textwidth]{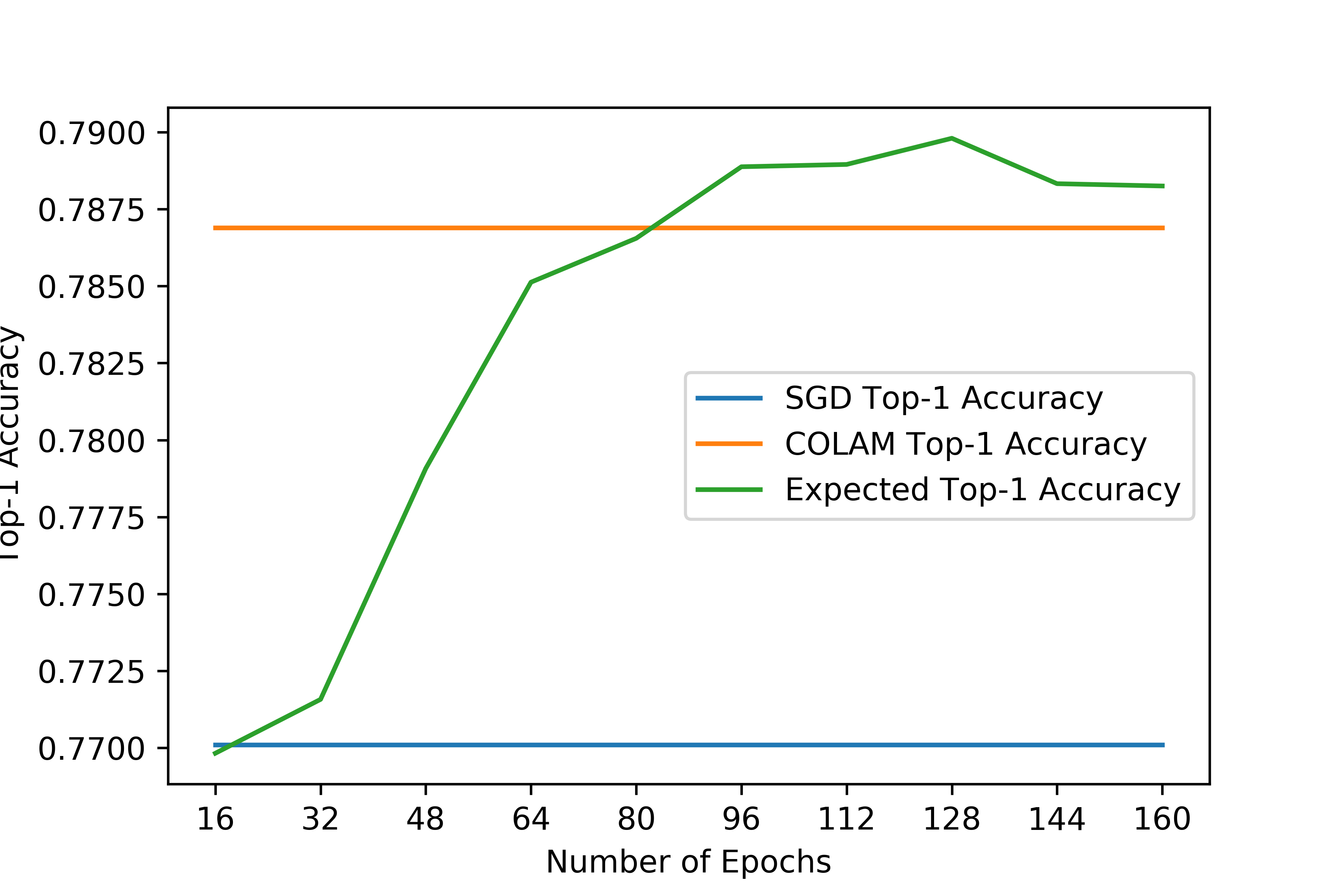}\label{expected_accu}}
\end{center}
   \caption{Demonstration of the effect of COLAM through learning curve and expected accuracy. Experiments are performed on PreActResNet34. SGD accuracy refers to the test accuracy of models that are trained with hard labels using a vanilla SGD optimizer. }
\end{figure*}

\subsection{Results}
\noindent\textbf{Image Classification} Table.~\ref{cifar10} and Table.~\ref{cifar100} shows that our COLAM consistently and significantly improve baseline models in accuracy for the majority of neural network architectures we tested. On CFIAR100, the improvement is generally within 1\%-2\% in comparison to models trained with hard labels using a vanilla SGD optimizer. In tasks of CIFAR10 and CIFAR100, we find that models with more complex architectures are not guaranteed to be better than simpler ones. For example, WideResNet40x4 with 9 million parameters outperforms PreActResNet34 with 20 million parameters. This happens regardless of the training technique used. We can observe a similar improvement in Table.~\ref{imagenet} when we apply COLAM to different models on ImageNet.

These results indicate the effectiveness of COLAM, which not only outperforms the traditional label smoothing technique, but it also beats other more dynamic but inherently equivalent form of label smoothing, namely Confidence Penalty and DisturbLabel. One reason that explains this phenomenon is that neither CP nor DisturbLabel encourages the model to learn the structural properties in the dataset when it regularizes the model. Preserving structural properties in the dataset is an important factor that contributes to good generalization of a model, as discussed in Sec.~\ref{di}.

\begin{table}
\caption{\textbf{Test accuracy using ResNet101 on various datasets.} All abbreviations follow the same rule as in Table.~\ref{cifar10}.}
\label{transferlearning}
\begin{center}
\begin{tabular}{P{2.6cm}P{1.4cm}P{1.4cm}P{1.4cm}}
\toprule
\textbf{Dataset}  & \textbf{HL} &\textbf{LS} & \textbf{COLAM} \\
\midrule
Flower102 & 0.9179 & 0.9279 &	\textbf{0.9294} \\
FGVC\_Aircraft & 0.7741	 & 0.7675 & \textbf{0.7787} \\
DTD &  0.6646 &	0.6705 & \textbf{0.6824} \\
CUB\_200\_2011 &  0.8172 &	0.8152 & \textbf{0.8246}\\
\bottomrule
\end{tabular}
\end{center}
\end{table}




\noindent\textbf{Transfer Learning} 
We fix our test model to be ResNet101 and perform experiments on the chosen datasets. Results in Table.~\ref{transferlearning} indicate that, compared with models trained using hard labels, COLAM improves the transfer learning outcomes on all four datasets, and the improvements range from 0.46\% to 1.78\%. These results testify that COLAM can enhance model performance on varying datasets.

In contrast, LS does not always yield positively improved results. Furthermore, the extent of improvement LS brings is considerably less than that of COLAM, as shown by the statistics.

\section{Discussion}

\subsection{Effect on Training Procedure}
To further demonstrate the effectiveness of our proposed alternating optimization of the training loss and soft labels, we dive into the training process of COLAM. Experiments show how the evolution of soft labels helps learning. 

We first plot the learning curve of the whole training procedure of PreActResNet34, as shown in Fig.~\ref{learningcurve}. For better demonstration purpose, we divide the training process into only 4 stages for COLAM training, with each stage consisting of 40 epochs. We observe that models trained with SGD using hard labels and COLAM display almost the same standard of performance in the first stage as expected. While in the 41st epoch, both training and test accuracy of COLAM get a sharp rise due to starting to involve soft labels generated in the 40th epoch. Then both training and test accuracy values drop slightly for a few epochs, and then return to the trend of slowly rising for the remaining epochs until next stage. A similar phenomenon also appears at the beginning of next stage, although the magnitude of accuracy improvement becomes much smaller. 
We notice that since the first sharp rising, COLAM continuously outperforms training using hard labels on test set by a stable gap for the remaining epochs until convergence. 

We do additional experiments to validate the effect of gradual promotion of soft labels, through our proposed alternative minimization approach. We divide the training process into 10 stages, each of which consists of 16 epochs. By performing COLAM, we obtain 10 checkpoints of soft labels. We train a model from scratch with vanilla SGD, except one difference that we use the supervision of these checkpoints. The top-1 accuracy of using each soft label is denoted as the corresponding ``expected accuracy". It is a solid measure of the quality of a soft label. As illustrated in Fig.~\ref{expected_accu}, we observe that the expected accuracy gradually increases with the evolution of soft labels. In detail, the expected accuracy grows fast during the early period of the whole training procedure. This observation verifies that the quality of soft labels is indeed improved by alternating optimized with the training loss. It is worth noting that the expected accuracy begins to surpass COLAM after about half of the training epochs. The expected accuracy shows a slight drop near the end of the training process, indicating that alternating optimization of the two objectives may suffer from over-fitting to a relatively small extent. Although COLAM improves the accuracy significantly, our experiments of the expected accuracy implies the potential existence of better design of soft labels.

\begin{figure*}[t]
\centering
  \subfloat[Training with HL]{\includegraphics[width=0.33\textwidth]{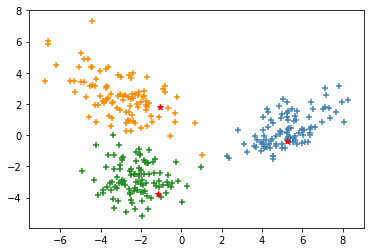}\label{train_sgd}}
  \subfloat[Test with HL]{\includegraphics[width=0.33\textwidth]{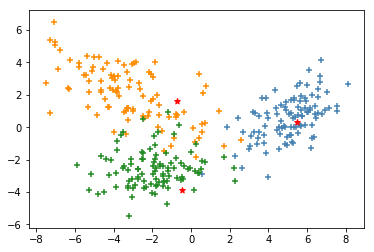}\label{val_sgd}}
  \subfloat[Training with LS]{\includegraphics[width=0.33\textwidth]{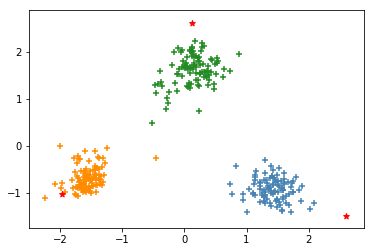}\label{train_ls}}\\
  \subfloat[Test with LS]{\includegraphics[width=0.33\textwidth]{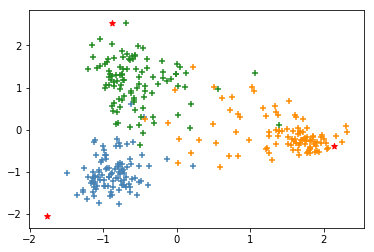}\label{val_ls}}
  \subfloat[Training with COLAM]{\includegraphics[width=0.33\textwidth]{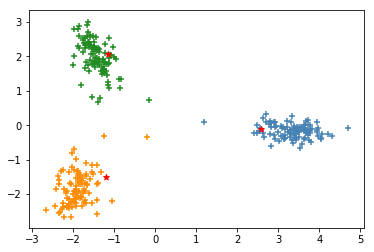}\label{train_COLAM}}
  \subfloat[Test with COLAM]{\includegraphics[width=0.33\textwidth]{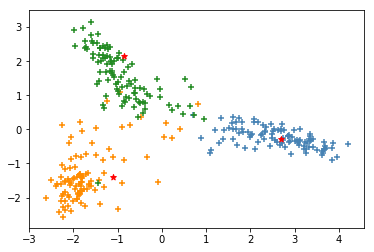}\label{val_COLAM}}
  
\caption{\textbf{Visualization of the penultimate layer representations and their relationships with the templates.} The blue, orange and green clusters correspond to ``man", ``palm tree", and ``pine tree" respectively. The red stars nearby are the templates of each cluster. ``HL" abbreviates ``hard labels."}

\label{vis}
\end{figure*}

\subsection{Effect on Deep Representations}
The empirical characteristics of COLAM observed in our experiments also show that COLAM is able to promote training by involving internal structural knowledge. We demonstrate this advantage through both qualitative and quantitative analysis.

\label{di}
\noindent\textbf{Qualitative Visualization} Recent work~\cite{whenMuller} gives insights into why label smoothing improves model performance. They argue that the logit $\pmb{x}^T\pmb{w}_k$ for class $k$ is correlated to the distance between $\pmb{x}$ and $\pmb{w}_k$, where $\pmb{x}$ is the penultimate layer representation and $\pmb{w}_k$ is the template for class $k$. As a result of their analysis, the penultimate layer representation $\pmb{x}$ should be close to the correct class template $\pmb{w}_k$ and equally distant from incorrect class templates $\pmb{w}_i$ for all $i\neq k$ after a model is trained with LS. 


Since our COLAM preserves the dataset structural properties compared to LS, $\pmb{x}'$ is expected to be closer to the class template that shares a greater extent of inter-class similarity than a class that does not. We verify this by projecting both the penultimate layer representation and template in 2D.
We choose 100 samples from each of three classes in CIFAR100 for this visualization: ``man", ``palm tree", and ``pine tree." Intuitively, ``palm tree", and ``pine tree" should be more similar to each other than ``man." 

Fig.~\ref{train_sgd} and Fig.~\ref{val_sgd} show the cluster distributions on the training set and test set when ResNet56 is trained with hard labels using vanilla SGD. We observe that the clusters are close to their respective templates. However, they are generally spread out and scattered. The clusters of ``palm tree" and ``pine tree" are relatively closer compared with ``man." This reflects the structural property within the dataset.

As seen in Fig.~\ref{train_ls} and Fig.~\ref{val_ls}, 
the clusters of label smoothing are tighter and easier to separate. The three clusters also try to be equally distant from the other classes' templates, resulting in a situation where clusters are drawn inward closer to the center of the subspace formed by the templates. 
However, the structural properties of the dataset is no longer preserved. ``palm tree" and ``pine tree" have the same distance as that between ``palm tree" and ``man."

Next, we see in Fig.~\ref{train_COLAM} and Fig.~\ref{val_COLAM} that, when ResNet56 is trained with COLAM, the clusters are better separated in both training set and test set. Each cluster remains close to its own template, but further away from the center of the subspace this time. This indicates the model's improved ability to distinguish each sample. Additionally, each cluster looks tighter in comparison to the clusters in Fig.~\ref{train_sgd} and Fig.~\ref{val_sgd}. What remains unchanged is the structural properties in the dataset:  ``palm tree" and ``pine tree" are still closer and ``man" is further away from these two. In comparison, LS does not maintain this structural property.

Observing all figures as a whole, we see that our method COLAM is a ``neutralizer" between using hard labels and LS. COLAM enjoys both accurately representing dataset structural properties (shown in training with hard labels) and obtaining tighter clusters that are easier for classification (shown in training with LS). This ``neutralizing" effect enables a model to better generalize.

\begin{figure*}[t]
\centering
  \subfloat[ADT (HL)]{\includegraphics[width=0.33\textwidth]{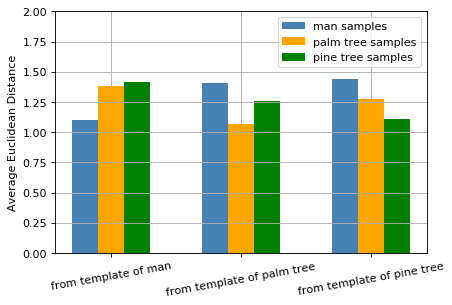}\label{temp_sgd}}
  \subfloat[ADT (LS)]{\includegraphics[width=0.33\textwidth]{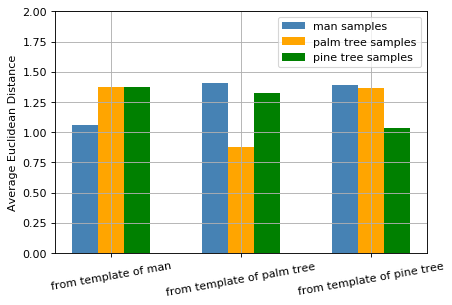}\label{temp_ls}}
  \subfloat[ADT (COLAM)]{\includegraphics[width=0.33\textwidth]{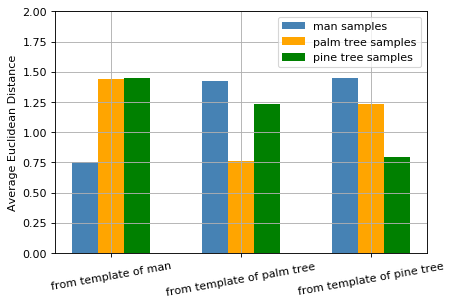}\label{temp_COLAM}}\\ 
  \subfloat[ADC (HL)]{\includegraphics[width=0.33\textwidth]{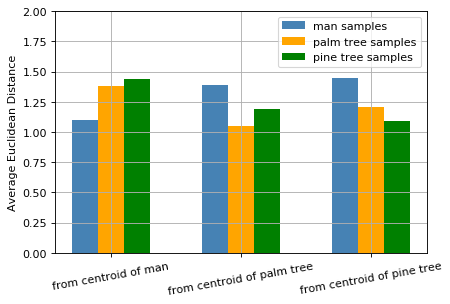}\label{centroid_sgd}}
  \subfloat[ADC (LS)]{\includegraphics[width=0.33\textwidth]{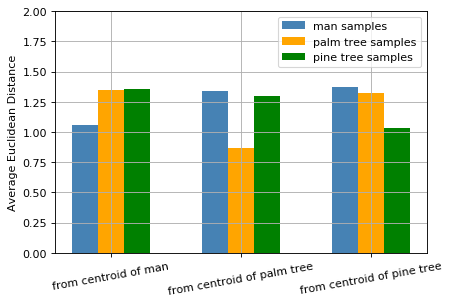}\label{centroid_ls}}
  \subfloat[ADC (COLAM)]{\includegraphics[width=0.33\textwidth]{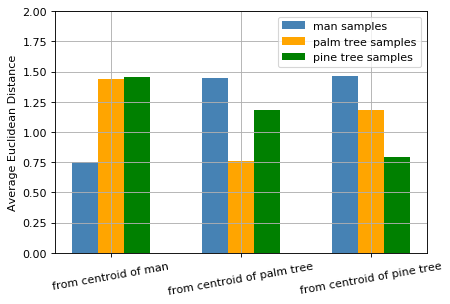}\label{centroid_COLAM}}
\caption{Clustering effects: \textbf{Average distance from each template (ADT)/centroid (ADC) of the cluster to samples.} Fig.~\ref{temp_sgd} to~\ref{temp_COLAM} show Average distance from each template to samples while Fig.~\ref{centroid_sgd} to~\ref{centroid_COLAM} show that from each centroid. ``HL" abbreviates ``hard labels". ``LS" abbreviates ``label smoothing". }
\label{dist_temp}
\end{figure*}

\noindent\textbf{Quantitative Evaluation}
We quantitatively evaluate three distance criteria for the same three classes to confirm the qualitative findings described above. Specifically, we find 
\begin{itemize}
    \item the Euclidean distance between the normalized templates, shown in Table~\ref{temp_dist}.
    \item the average Euclidean distance between the template and samples' penultimate layer representations, shown in Fig.~\ref{temp_sgd} to Fig.~\ref{temp_COLAM}.
    \item the average Euclidean distance between the centroid and samples' penultimate layer representations, shown in Fig.~\ref{centroid_sgd} to Fig.~\ref{centroid_COLAM}.
\end{itemize}

Note that we normalize all vectors when we compute such distances in order to make fair comparisons.

As shown in Table~\ref{temp_dist}, COLAM largely preserves the structural properties in the dataset by keeping the distance between ``palm tree", and ``pine tree" templates closer and distance from ``man" greater, which is consistent with the model trained with hard labels. 

In contrast, LS generally enlarges the overall distance in between templates, due to the requirement that each cluster should be equi-distant away from the incorrect class templates. Since clusters will stick close to their respective templates after training, this observation implies that data structural properties is missing in LS. 



Fig.~\ref{temp_sgd} to Fig.~\ref{temp_COLAM} shows the distance between templates and clusters. 
We see that COLAM (Fig.\ref{temp_COLAM}) is able to preserve structural properties in the dataset because the distance from a template to other clusters display the same trend as that in ~\ref{temp_sgd}. Together, the two figures also show that each template is closer to its corresponding cluster and further away from other clusters when the model is trained with COLAM. Since Table.~\ref{temp_dist} shows that the distance between the templates in these two methods are roughly the same, the difference between \emph{the average distance from a template to its own cluster} and \emph{the average distance from a template to other clusters} indicates how well the clusters are separated. Fig.\ref{temp_COLAM} shows a larger such difference than Fig.\ref{temp_sgd} does. This explains why COLAM outperforms training with hard labels.

\begin{table}
\caption{\textbf{Euclidean distance between the normalized templates.} Same abbreviations are used as found in Table.~\ref{cifar10}.}
\label{temp_dist}
\begin{center}
\begin{tabular}{P{3.2cm}P{1.5cm}P{1.5cm}P{1.5cm}}
\toprule
\textbf{Distance}  & \textbf{HL} &\textbf{LS} & \textbf{COLAM} \\
\midrule
(Man, Palm Tree) & 1.433     &1.460 & 	1.438 \\
(Man, Pine Tree) & 1.425     &  1.400 & 1.461 \\
(Palm Tree, Pine Tree) & 1.264   & 1.369 & 1.227\\
Average & 1.374  & 1.410 & 1.375\\
\bottomrule
\end{tabular}
\end{center}
\end{table}

\begin{figure*}
\begin{center}
   \subfloat[\# peer samples vs. \# epochs per stage ]{\includegraphics[width=0.35\textwidth]{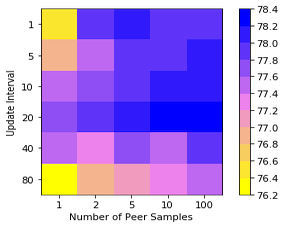}\label{hyper}}
   \subfloat[Sensitivity of $T$]{\includegraphics[width=0.41\textwidth]{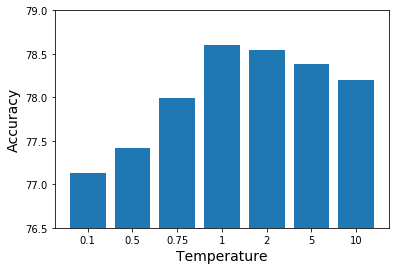}\label{overall}}
\end{center}
   \caption{COLAM hyperparameter experiments on PreActResNet18.}
\end{figure*}

Now we compare COLAM (Fig.~\ref{temp_COLAM}) with LS (Fig.~\ref{temp_ls}). Because COLAM does not need to force each cluster to be equi-distant away from the incorrect class templates and preserves structural properties of the data, the distance between a template to its own cluster can get far smaller than that in LS. This smaller distance contributes to a larger difference between \emph{the average distance from a template to its own cluster} and \emph{the average distance from a template to other clusters}. Hence, COLAM is able to improve model generalization to a greater extent than LS. Moreover, clusters in LS are almost equi-distant away from other classes' templates as shown in Fig.~\ref{temp_ls}, this violates the structural properties in the dataset.

Computing the average distance between samples and their centroid reveals how tight a cluster is. A smaller average value indicates a tighter cluster and vice versa. In Fig.~\ref{centroid_sgd} to Fig.~\ref{centroid_COLAM}, we observe the smallest such distance is found in the model trained with COLAM. 

We are also interested in the difference between \emph{the distance from a centroid to its own cluster} and \emph{the distance from a centroid to other clusters}. The larger this difference is, the better the clusters are separated. Fig.~\ref{centroid_COLAM} indicates that the model trained with COLAM yields the greatest such distance. This, from a slightly different view, explains why COLAM gives rise to the highest generalization ability.


\subsection{Choice of Hyperparameters}

The most important two hyperparameters are the number of stages and number of random peer samples. We use update intervals, or equivalently number of epochs per stage, instead of number of stages for clarity in this experiments. We set update intervals to be $[1, 5, 10, 20, 40, 80]$ and number of peer samples to be $[1, 2, 5, 10, 100]$. We run a grid search method to validate different combinations of these two variables.

In Fig.~\ref{hyper} we see that performance of COLAM does not seem to be very sensitive to most combinations of the hyperparameters. When the number of peer samples increases to 5 or more, model accuracy tends to be over 77.3\%. Even when the number of peer samples is low, a good choice of the value of update interval can boost the model performance significantly. Low accuracy of the model only happens consistently when the value of update interval is large.

Another important hyperparameter is the temperature $T$, which softens the probability distribution of incorrect classes of soft labels. We explore $T=[0.1, 0.5, 0.75, 1, 2, 5, 10]$ used in PreActResNet34 on CIFAR100. 
Theoretically, a larger value of $T$ makes the probability distribution of the incorrect classes smoother. When $T$ gets sufficiently large, 
this will make our COLAM to behave like traditional LS. In contrast, the probability distribution among incorrect classes gets even sharper if $T$ is less than 1. When $T$ gets close to 0, it is closed to the original hard label.
Empirically we recommend $T$ to be some value between 1 to 2.

\section{Conclusion}
In this paper, we have discussed the advantages of using soft labels as the target in deep learning and proposed a novel algorithm COLAM that alternatively minimizes the training loss subject to the soft label and the objective to learn improved soft labels. We have conducted numerous experiments to demonstrate the method's effectiveness on a variety of tasks. We have also offered both qualitative and quantitative explanations as to why COLAM is more beneficial than existing techniques to produce soft labels.

\bibliographystyle{unsrt}  
\bibliography{sample-base}

\end{document}